\documentclass[runningheads]{llncs}
\usepackage{graphicx}
\usepackage{caption,subcaption}
\usepackage{booktabs}
\usepackage[font=small]{caption}
\usepackage{tikz}
\usepackage{xfrac}
\usepackage{hyperref}
\hypersetup{
    colorlinks=true,
    linkcolor=red,
    filecolor=magenta,      
    urlcolor=cyan,
    pdftitle={Overleaf Example},
    pdfpagemode=FullScreen,
    }

\newcommand\blfootnote[1]{%
  \begingroup
  \renewcommand\thefootnote{}\footnote{#1}%
  \addtocounter{footnote}{-1}%
  \endgroup
}

\usepackage{paralist} 
\newcommand{\etal}{\textit{et al.}}
\usepackage{comment}
\usepackage{amsmath,amssymb} 
\usepackage{xcolor,colortbl}
\definecolor{shadecolor}{RGB}{255,203,203}
\definecolor{lightgreen}{RGB}{39,179,118}
\usepackage{paralist} 
\usepackage{cite}
\usepackage{wrapfig}
\usepackage[accsupp]{axessibility}  

\definecolor{redred}{RGB}{196,23,63}
\definecolor{pinkpink}{RGB}{243,157,153}
\definecolor{greengreen}{RGB}{200,228,206}
\definecolor{blueblue}{RGB}{168,218,224}
\definecolor{orangeorange}{RGB}{231,96,83}
\definecolor{graygray}{RGB}{229,229,229}
\usepackage{colortbl}

\begin{document}
\pagestyle{headings}
\mainmatter
\def\ECCVSubNumber{29}  

\title{Students taught by multimodal teachers are superior action recognizers} 

\titlerunning{Students taught by multimodal teachers are superior action recognizers}
%
\author{Gorjan Radevski\textsuperscript{*} \and Dusan Grujicic\textsuperscript{*} \and Matthew Blaschko \and Marie-Francine Moens \and Tinne Tuytelaars}
\authorrunning{G. Radevski and D. Grujicic, et al.}
%
\institute{KU Leuven University, Belgium\\
\email{firstname.lastname@kuleuven.be}}
\maketitle

\begin{abstract}
The focal point of egocentric video understanding is modelling hand-object interactions. Standard models -- CNNs, Vision Transformers, etc. -- which receive RGB frames as input perform well, however, their performance improves further by employing \textit{additional} modalities such as object detections, optical flow, audio, etc. as input. The added complexity of the required modality-specific modules, on the other hand, makes these models impractical for deployment.
The goal of this work is to retain the performance of such multimodal approaches, while using \textit{only} the RGB images as input at inference time. 
Our approach is based on multimodal knowledge distillation, featuring a multimodal teacher (in the current experiments trained only using object detections, optical flow and RGB frames) and a unimodal student (using only RGB frames as input). We present preliminary results which demonstrate that the resulting model -- distilled from a multimodal teacher -- significantly outperforms the baseline RGB model (trained without knowledge distillation), as well as an omnivorous version of itself (trained on all modalities jointly), in both standard and compositional action recognition.\blfootnote{\textsuperscript{*}Authors contributed equally}
\end{abstract}

\section{Introduction \& Background}
Various egocentric action recognition methods \cite{radevski2021revisiting, herzig2022object, zhang2022object, materzynska2020something, wang2018videos, yan2020interactive, kim2021motion} demonstrate that explicitly modelling hand-object interactions (usually represented via bounding boxes \& object categories) significantly improves the action recognition performance, especially notable in compositional and few-shot setups \cite{radevski2021revisiting}. Similarly, other works show that leveraging multiple modalities at inference time yields improved performance \cite{xiong2022m, gabeur2020multi, kazakos2021little, nagrani2021attention}. The assumption these methods make is that all modalities used during training, are also available at inference time, thus making them impractical or even impossible to use in practice, e.g., on a limited compute budget such as embedded devices.
Namely, using dedicated models for each additional modality (e.g., object detector + tracker + transformer when using bounding boxes \& object categories \cite{radevski2021revisiting}), increases both the memory footprint as well as the inference time. 
\textit{Ideally, we leverage additional modalities during training, while the resulting model uses only RGB frames at inference time, e.g., when deployed in practice.}\par
One way to achieve the aforementioned goal is training omnivorous models, i.e., models trained jointly on multiple modalities, which have been shown to generalize better \cite{girdhar2022omnivore}. In this work, we take on a different route to transfer multimodal knowledge to models eventually used in a unimodal setting. Namely, we attempt to distill the knowledge from a multimodal ensemble -- exhibiting superior performance, but impractical to be used in practice -- to a standard RGB-based action recognition model \cite{bertasius2021space}. By employing state-of-the-art knowledge distillation practices \cite{beyer2022knowledge}, we show that such student, taught by a multimodal teacher, significantly outperforms its unimodal baseline counterpart (trained without knowledge distillation), as well as an omnivorous version of itself (trained on all modalities jointly). Code and trained models will be released upon publication of the full paper.

\section{Methods}
We train a separate model for each of the modalities we use, namely RGB frames, optical flow and bounding boxes and categories of the scene objects. For the default -- RGB frames modality -- we use TimesFormer \cite{bertasius2021space} (ViT-B variant), pre-trained on ImageNet. For optical flow, we use the same pre-trained TimesFormer model, which we directly apply on optical flow frames. For bounding boxes \& categories, we use STLT \cite{radevski2021revisiting}. Our multimodal teacher simply averages the logits of all modality-specific models. For the student, we use TimesFormer, which is given only the RGB frames as input -- which we use at inference time to measure how much of the multimodal teacher performance we retain.\par
During training we perform multimodal knowledge distillation, where, as originally proposed by Hinton \etal \cite{hinton2015distilling}, we minimize the KL-divergence between the class probabilities predicted by the teacher $\mathbf{p}^t = [p^t_1, p^t_2, \dots, p^t_{C}] \in \mathbb{R}_+^C$, and the student $\mathbf{p}^s = [p^s_1, p^s_2, \dots, p^s_{C}] \in \mathbb{R}_+^C$:

\begin{equation}
\begin{gathered}
    \text{KL}(\mathbf{p}^t || \mathbf{p}^s) =  \sum_{i=1}^{C} \left( - p^t_i\log p^s_i + p^t_i\log p^t_i \right),
\end{gathered}
\end{equation}

where $C$ is the number of classes. Additionally, we use a temperature parameter $\tau$ to control the entropy of the probability scores while preserving their ranking $p_i \propto \text{exp}(\frac{\log p_i}{\tau})$. During training we follow the patient \& consistent teaching paradigm \cite{beyer2022knowledge}, i.e., the student and the teacher strictly receive the same views of the data, which in our use-case are temporal and spatial crops.

\begin{figure}[t]
\centering
\includegraphics[width=0.8\textwidth]{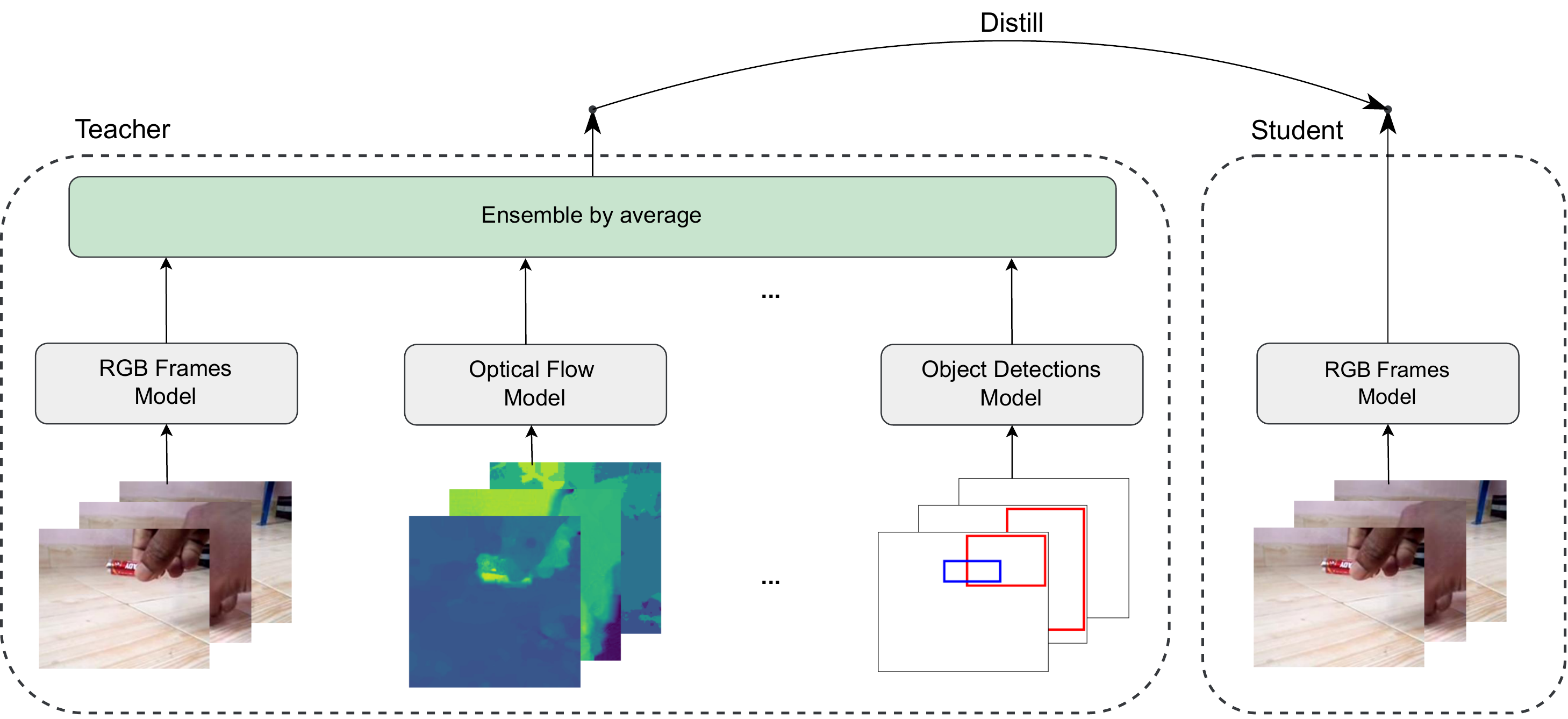}
\caption{Multimodal teacher, employing RGB frames, optical flow \& object detections,; Unimodal student, taught \textit{only} using RGB frames.}
\label{fig:mainfigure}
\end{figure}
\section{Experiments \& Discussion}\label{sec:experiments}

\textbf{Datasets.} Currently, we perform experiments on the Something-Something \cite{goyal2017something} (V2) and the Something-Else \cite{materzynska2020something} (compositional setup) datasets. Something-Something is an egocentric video dataset of people performing actions with their hands, with 174 unique object agnostic actions, e.g., ``pushing [something] left''. Nevertheless, the objects the people interact with might overlap between training and testing, indicating that RGB frames based models might pick up undesireable biases by overfitting on the objects' appearance to discriminate between the actions. Therefore, Something-Else proposes a compositional generalization data split, on which the performance of standard video models deteriorates. Here, the objects at training and testing time do not overlap, and hence, the models encounter strictly novel objects during testing (the number of actions remains the same, i.e., 174 unique actions). In future work we would also validate the performance improvements on a few-shot version of Something-Else, as well as the Epic-Kitchens 55/100 \cite{damen2018scaling, damen2020rescaling} datasets -- including long-tail actions and kitchen environments unseen during training.\par

\textbf{Experimental setup.} We train all models for 20 epochs using AdamW \cite{loshchilov2017decoupled}, a peak learning rate of $5e-5$, linearly increased for the first 10\% of training and decreased to 0.0 till the end, weight decay of $1e-3$, gradient clipping of 5.0, and temperature $\tau$ of 10.0. We randomly sample 8 frames during training and apply random spatial frame cropping and color jitter, while during inference, we take frames at regular intervals without any test-time augmentations (e.g., multiple spatial and temporal crops) to keep the setup simple. 

\textbf{Modalities.} We use optical flow, released by \cite{wang2016temporal} for Something-Something and Something-Else, and bounding boxes \& categories released by \cite{herzig2022object} for Something-Something and \cite{materzynska2020something} for Something-Else.

\textbf{Competing models.} (i) Baselines, trained individually on each modality; (2) Teachers, ensembles by logits average of baselines trained on individual modalities; (iii) Omnivore, trained on \textit{all modalities} jointly by randomly sampling a modality for each video during training. Object detections + categories represented as 2px colored bounding boxes pasted on a white canvas, e.g., hand in blue and object in red. Trained 3x longer than the other models to account for the random sampling of modalities during training.

\textbf{Discussion.} In Table~\ref{table:smth-smth} we report performance on the Something-Something and the Something-Else (compositional setting) datasets. As reported by others \cite{radevski2021revisiting,materzynska2020something}, we observe that the RGB trained baseline yields significantly lower performance on compositional data, while the baselines trained on other modalities (obj. detections \& optical flow) are more robust. Similarly, as reported by other works \cite{xiong2022m, radevski2021revisiting, materzynska2020something}, leveraging additional modalities at inference time improves performance, albeit at the cost of higher memory and time complexity.\par
A novel observation in our work is that multimodal knowledge distillation works well in the context of egocentric video understanding, and sometimes the student even outperforms the teacher (when the teacher gets RGB frames \& obj. detections as input). Further, when employing all modalities (RGB frames, obj. detections \& optical flow), the student performance improves further, and outperforms the baseline by \textbf{\textcolor{redred}{7.7\%}} as per top-1 accuracy in the compositional setup. Lastly, as reported by Girdhar \etal \cite{girdhar2022omnivore}, the omnivorous model is superior to the baseline model, however, it is outperformed by the student taught by a multimodal teacher. This suggests that multimodal knowledge transfer to a unimodal model is superior when distilling from a previously taught multimodal teacher, compared to joint training with all modalities.

\begin{table}[t]
\centering
\resizebox{0.9\textwidth}{!}{
\begin{tabular}{>{\columncolor{graygray}}c>{\columncolor{pinkpink}}c>{\columncolor{blueblue}}c>{\columncolor{blueblue}}c>{\columncolor{greengreen}}c>{\columncolor{greengreen}}c} \toprule
    {} & {} & \multicolumn{2}{>{\columncolor{blueblue}}c}{Something-Something} & \multicolumn{2}{>{\columncolor{greengreen}}c}{Something-Else} \\
    {Method} & {Modalities at inference} & {Top 1} & {Top 5} & {Top 1} & {Top 5} \\ \midrule
    Baseline \cite{bertasius2021space} & RGB frames & 59.6 & 85.6 & 51.7 &
    78.1 \\
    Baseline \cite{radevski2021revisiting} & Obj. detections & 44.2 & 73.6 & 39.5 & 66.3 \\
    Baseline \cite{bertasius2021space} & Optical flow & 50.2 & 79.6 & 49.7 & 78.3 \\ \midrule
    Teacher \cite{bertasius2021space, radevski2021revisiting} & RGB Frames \& Obj. detections & 62.6 & 87.6 & 57.2 & 82.7 \\
    Student & RGB frames & 62.1\textsubscript{\textcolor{redred}{\textbf{+2.5}}} & 88.0\textsubscript{\textcolor{redred}{\textbf{+2.4}}} & 57.2\textsubscript{\textcolor{redred}{\textbf{+5.5}}} & 83.5\textsubscript{\textcolor{redred}{\textbf{+5.4}}} \\ \midrule
    Teacher \cite{bertasius2021space} & RGB Frames \& Optical flow & 64.0 & 89.0 & 61.5 & 86.2 \\
    Student & RGB frames & 62.3\textsubscript{\textcolor{redred}{\textbf{+2.7}}} & 88.8\textsubscript{\textcolor{redred}{\textbf{+3.2}}} & 57.6\textsubscript{\textcolor{redred}{\textbf{+5.9}}} & 84.1\textsubscript{\textcolor{redred}{\textbf{+6.0}}} \\ \midrule
    Teacher \cite{bertasius2021space, radevski2021revisiting} & RGB Frames \& Optical flow \& Obj. detections & 65.7 & 90.2 & 63.2 & 87.3 \\
    Omnivore \cite{bertasius2021space, girdhar2022omnivore} & RGB frames & 62.5 & 88.1 & 56.8 & 83.3 \\
    Student & RGB frames & 63.9\textsubscript{\textcolor{redred}{\textbf{+4.3}}} & 89.2\textsubscript{\textcolor{redred}{\textbf{+3.6}}} & 59.4\textsubscript{\textcolor{redred}{\textbf{+7.7}}} & 85.4\textsubscript{\textcolor{redred}{\textbf{+7.3}}} \\
    \bottomrule
\end{tabular}
}
\caption{Something-Something \cite{goyal2017something} \& Something-Else \cite{materzynska2020something} (compositional split) action recognition accuracy. Improvement over RGB frames baseline \cite{bertasius2021space} in \textcolor{redred}{\textbf{red}}.}
\label{table:smth-smth}
\end{table}
\section{Conclusion \& Future work}
In the current version of this work, we observe that unimodal student models distilled from multimodal teachers outperform their non-distilled, and even omnivorous counterparts, on regular and compositional egocentric action recognition. We consider this observation as valuable since multimodal action recognition models are unfeasible to use in practice, while we show that, using multimodal knowledge distillation, a unimodal action recognition model yields performance competitive with such multimodal models.\par

\textbf{Future work.} The next steps involve verifying: (i) the extent to which the benefits of the proposed approach are consistent across different model sizes, e.g., ViT-L (Girdhar \etal \cite{girdhar2022omnivore} observe higher performance gains with larger omnivorous models); (ii) robustness against covariate shifts; (iii) benefits of distilling from additional modalities, e.g., audio; (iv) ability to cope with noun-verb compositions which occur rarely during training.
Regarding the latter three points, we will analyze the performance on Epic-Kitchens 55/100 datasets, as their unseen environments splits emphasize the discrepancy between the train and test visual environments (ii), they feature audio as an additional modality (iii), and are characterized by long-tailed distributions of actions, e.g., ''mix mushrooms`` occurs less frequently during training compared to other compositions (iv).
\section*{Acknowledgements}
We acknowledge funding from the Flemish Government under the Onderzoeksprogramma Artifici\"{e}le Intelligentie (AI) Vlaanderen programme.

\bibliographystyle{splncs04}
\bibliography{egbib}
\end{document}